\documentclass[letterpaper, 10 pt, conference]{ieeeconf}

\IEEEoverridecommandlockouts
\usepackage{amssymb}
\usepackage{newunicodechar}
\usepackage{booktabs}
\usepackage{graphicx}
\usepackage{array}
\usepackage{amsmath}
\usepackage{url}
\usepackage{makecell}
\usepackage{caption}
\usepackage{cite}
\usepackage{hyperref}

\hypersetup{
  colorlinks=true,
  citecolor=blue,
  linkcolor=blue,
  urlcolor=blue,
}

\makeatletter
\let\old@maketitle\@maketitle
\renewcommand{\@maketitle}{\old@maketitle
    \centering
    \vspace{5pt} 
    \includegraphics[width=\textwidth]{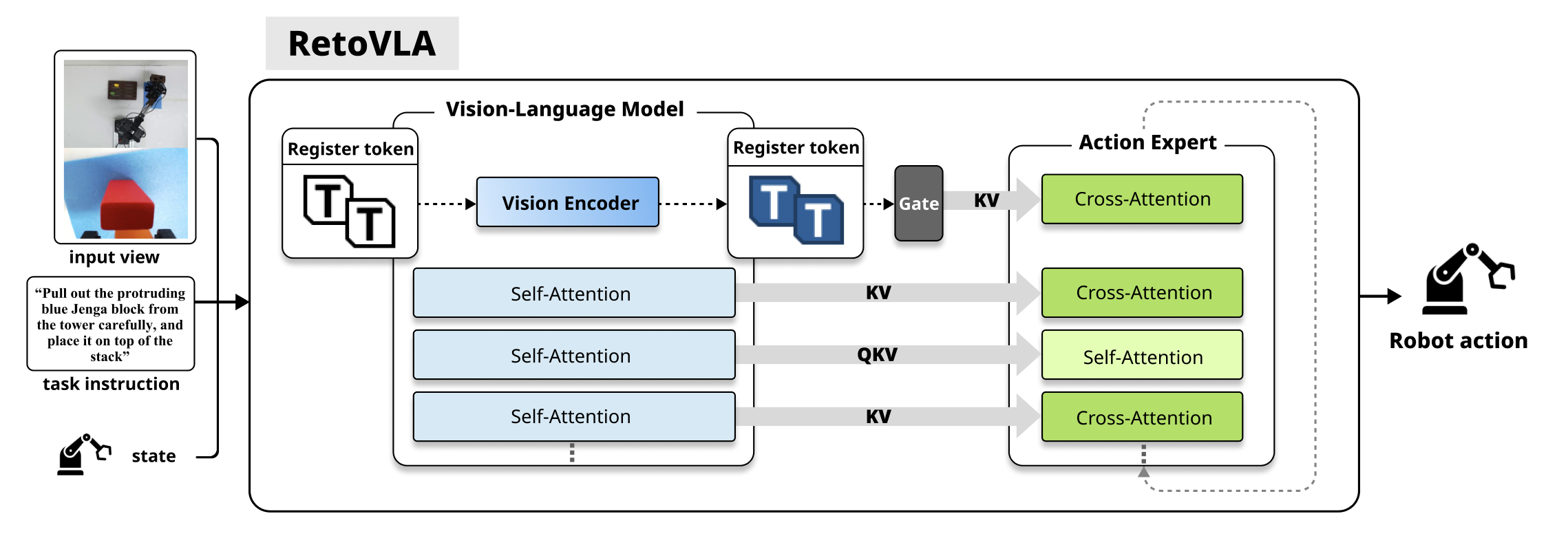} 
    \vspace{0.2em} 
    \captionof{figure}{Overview of the RetoVLA architecture. A dedicated spatial pathway (dashed arrow) injects global scene context by routing Register Tokens directly into the Action Expert. Unlike standard encoders that discard these tokens post-processing, RetoVLA repurposes them to seamlessly integrate spatial and semantic features, requiring no additional parameters.}
    \label{fig:architecture}
    \vspace{-10pt} 
}
\makeatother

\title{\LARGE \bf
RetoVLA: Reusing Register Tokens for Spatial Reasoning in Vision-Language-Action Models
}

\author{\textbf{Jiyeon Koo$^{1,\dagger}$, Taewan Cho$^{1,\dagger}$, Hyunjoon Kang$^{1}$, Eunseom Pyo$^{1}$, Tae Gyun Oh$^{1}$, Taeryang Kim$^{1}$,} \\ \textbf{and Andrew Jaeyong Choi$^{1,*}$}
\thanks{$^{\dagger}$These authors contributed equally to this work.}%
\thanks{$^{\dagger}$First authors: {\tt\small \{halo1225, taewan2002\}@gachon.ac.kr}.}
\thanks{$^{*}$Corresponding author: {\tt\small andrewjchoi@gachon.ac.kr}.}
\\ 
\textit{$^{1}$School of Computing, Gachon University}
}

\begin{document}

\maketitle
\thispagestyle{empty}
\pagestyle{empty}
\setcounter{figure}{1}


\begin{abstract}
Vision-Language-Action (VLA) models have demonstrated robust performance across diverse robotic tasks. However, their high memory and computational demands often limit real-time deployment. While existing model compression techniques reduce the parameter footprint, they often drop in 3D spatial reasoning and scene layout understanding. This work introduces RetoVLA, an architecture designed to maintain spatial awareness in lightweight models by repurposing Register Tokens—learnable parameters originally introduced to mitigate attention artifacts in Vision Transformers. While these tokens are generally discarded once used, we repurpose them for their dense representation of global spatial context. RetoVLA integrates these recycled tokens directly into the action-planning module through a dedicated spatial context injection path. Our proposed design enables the recovery of global context without increasing the total parameter count. Real-world experiments using a 7-DOF manipulator show a 17.1\%p improvement in average success rates over the baseline. Our results demonstrate that leveraging internal register tokens provides a highly effective mechanism for developing efficient, spatially-aware robotic agents. A video demonstration is available at: \url{https://youtu.be/2CseBR-snZg}
\end{abstract}
\section{INTRODUCTION}

Vision-Language-Action (VLA) models such as RT-2 \cite{zitkovich2023rt} and OpenVLA \cite{kim2024openvla} map natural language instructions to robotic motor commands. Through web-scale pre-training, they enable strong zero-shot generalization in unseen environments. However, their scale and computational cost remain a major bottleneck for real-time deployment on physical hardware.

\begin{figure*}[t!] 
    \centering
    \includegraphics[width=\textwidth]{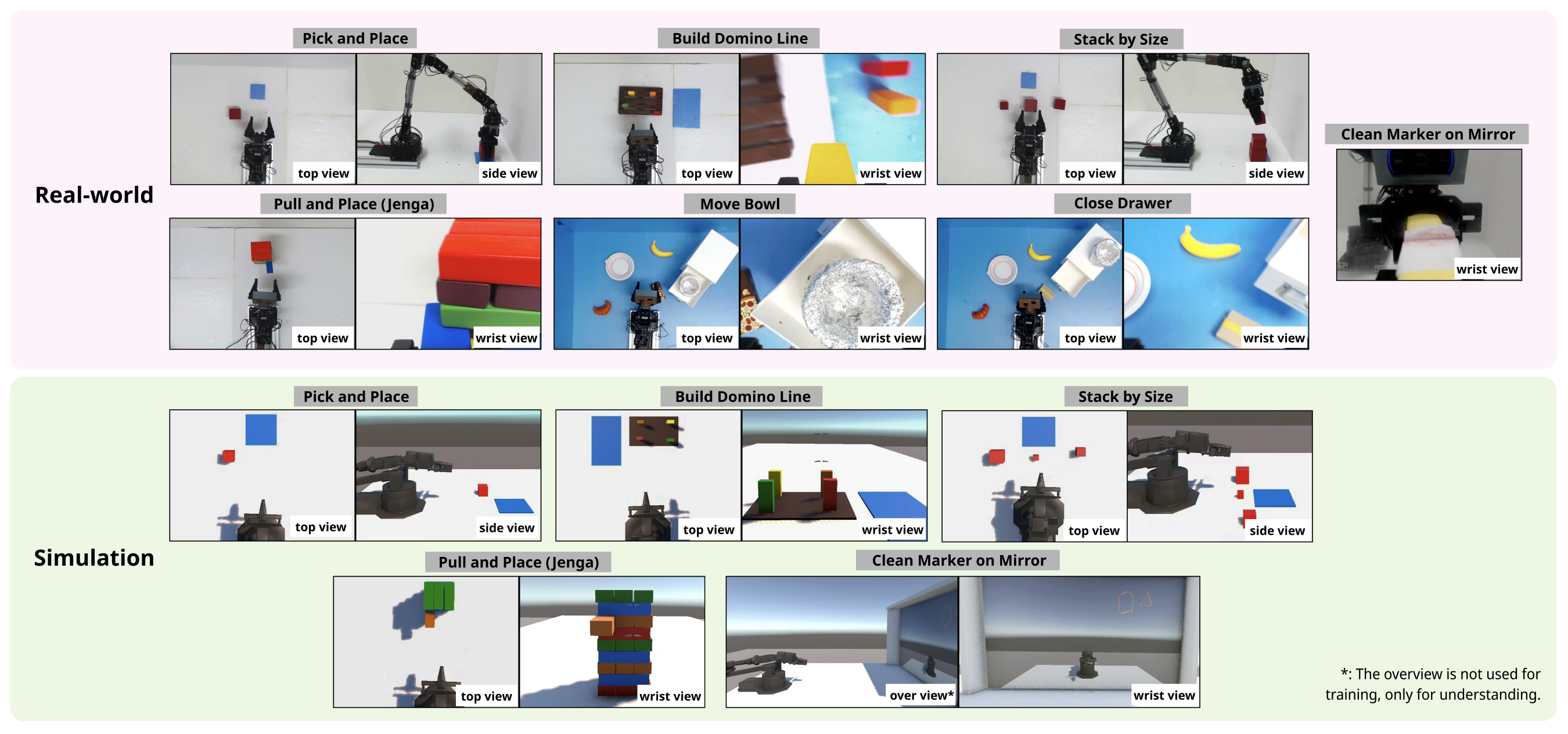} 
    \caption{Experimental setup overview. (Top) We use a custom robot arm for seven real-world manipulation tasks. Two tasks, `Move Bowl' and `Close Drawer', come directly from the LIBERO benchmark. (Bottom) We also implemented five additional tasks in the simulation to match our real-world setup, alongside the two LIBERO tasks.}
    \label{fig:tasks}
\end{figure*}

To address this efficiency issue, prior work has focused on smaller models such as SmolVLA \cite{shukor2025smolvla}. However, reducing model size comes at a cost. Lightweight models often lose the capacity to represent 3D layouts and spatial relationships.

We address this limitation by reusing information that is typically discarded. Darcet et al.~{\cite{darcet2023vision}} observed that large Vision Transformers (ViTs) {\cite{dosovitskiy2020image}}, such as DINOv2 {\cite{oquab2023dinov2}}, temporarily store global scene information in background image patches during training. Although this behavior supports global understanding, it degrades the visual details of those patches.

\begin{figure}[b!]
    \centering
    \includegraphics[width=\columnwidth]{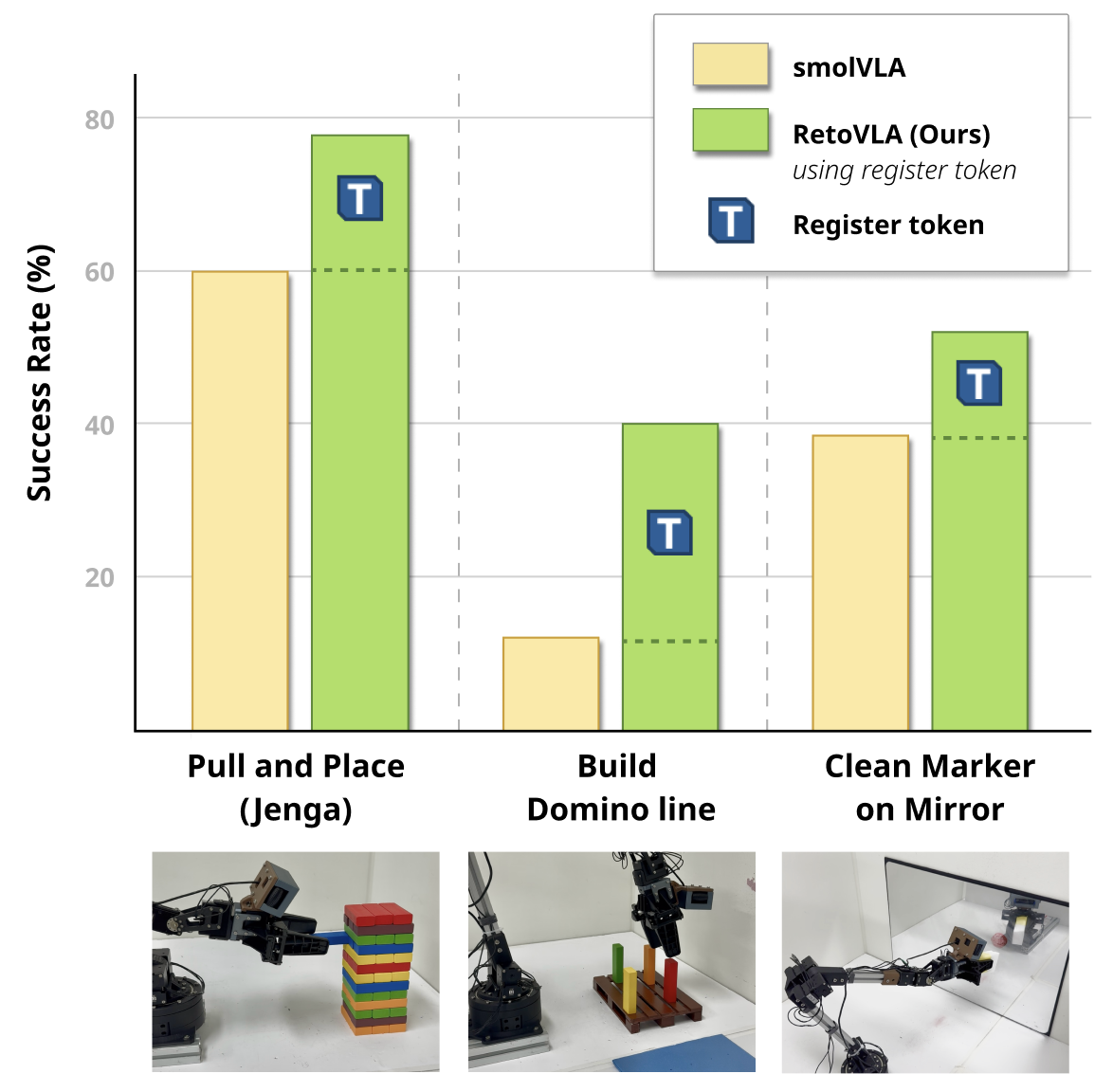}
    \vspace{-0.5em}
    \caption{Comparison of RetoVLA and the SmolVLA baseline on challenging real-world tasks. (Top) RetoVLA (green) significantly outperforms the baseline (yellow). (Bottom) This gain comes from reusing Register Tokens \cite{darcet2023vision} (indicated by the `T' icon) to inject global spatial context into the Action Expert.}
    \label{fig:main_fig}
\end{figure}

To mitigate these artifacts, researchers introduced Register Tokens \cite{darcet2023vision}. These tokens act as dedicated scratchpads that absorb global information while preserving the visual fidelity of image patches. Although they are usually discarded after use, we examine whether they retain meaningful spatial information.

\begin{figure*}[t!]
    \centering
    \includegraphics[width=\textwidth]{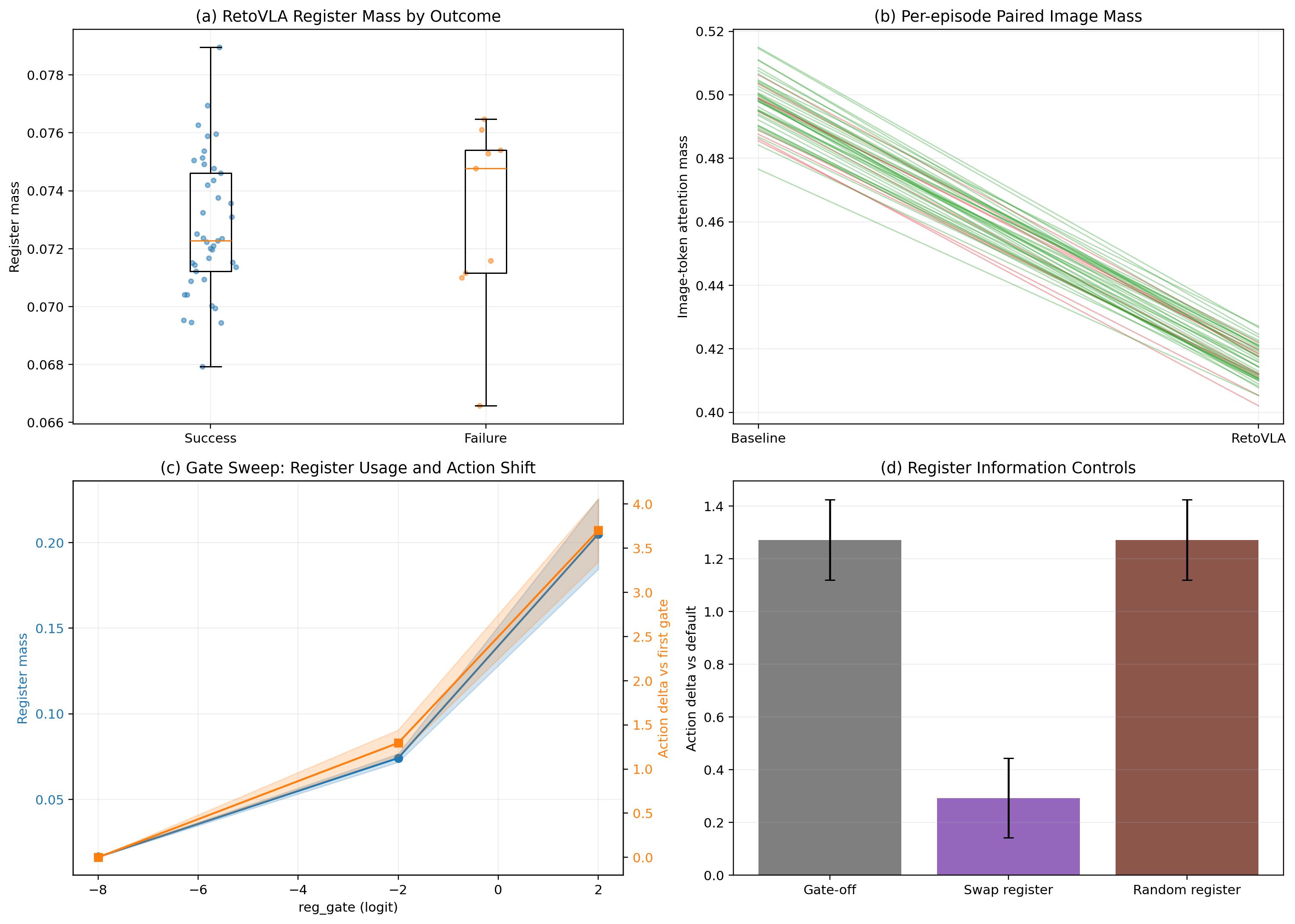}
    \caption{Causal Analysis of Register Tokens. (a) Consistent attention mass across outcomes indicates active utilization. (b) RetoVLA reduces attention on image patches, moving its attention to global context. (c) Gate value ($g$) directly changes the action output, establishing causality. (d) Randomized tokens degrade performance, confirming specific information encoding.}
    \label{fig:register_analysis}
\end{figure*}

We hypothesize that these tokens encode a highly compressed summary of workspace layouts and 3D relationships, and that preserving them can improve robotic scene understanding.

Based on this hypothesis, we propose RetoVLA (Reusing Register Tokens {\cite{darcet2023vision}} VLA). RetoVLA improves efficiency by recycling this latent information. Figure~\ref{fig:main_fig} summarizes the main result. Our contributions are:

1) A spatial context injection method: As shown in Figure~\ref{fig:architecture}, we repurpose Register Tokens {\cite{darcet2023vision}} from artifact absorbers into providers of spatial context and feed them directly into the Action Expert.

2) An efficient design: We show that these tokens recover spatial awareness lost in lightweight models such as SmolVLA {\cite{shukor2025smolvla}} without adding computational overhead.

3) Evaluation in simulation and on hardware: On the LIBERO benchmark and on a real 7-DOF robot, RetoVLA significantly outperforms the baseline, improving the real-world average success rate from 50.3\% to 67.4\% (+17.1\%p).

\section{RELATED WORK}

\subsection{Vision-Language-Action (VLA) Models}

Transformers are now the standard backbone for robot learning. Early systems such as RT-1 \cite{brohan2022rt} supported multi-task manipulation, while subsequent models connected large VLMs directly to robot actions. RT-2 \cite{zitkovich2023rt}, for example, represented robot actions as discrete text tokens, and more recent open-weight models such as OpenVLA \cite{kim2024openvla} and $\pi$0 \cite{black2024pi_0} show strong zero-shot generalization.

However, their billions of parameters lead to slow inference and create major bottlenecks for reactive physical robots. Recent work improves spatial awareness \cite{cai2025spatialbot}, visual robustness \cite{hancock2025run}, and low-data learning \cite{tang2025kalie, tong2024quart}, but the core problem of computational efficiency remains largely unresolved.

\subsection{Lightweight Vision-Language-Action Models}

To deploy VLAs on robots, recent work has developed smaller models. SmolVLA {\cite{shukor2025smolvla}} uses smaller backbones, layer skipping, and LoRA~\cite{hu2022lora} to reduce memory cost. Other methods explore model merging~\cite{cho2025research} or diffusion-based decoders~\cite{wen2025tinyvla, chi2023diffusion} for faster action generation.

Although these models are faster, they often lose critical 3D spatial reasoning and global context awareness. Prior work has tried to recover spatial information with external depth encoders~\cite{cho2025enhancing}, but those methods add computational overhead. RetoVLA takes a different approach: instead of adding new encoders, we reuse standard ViT Register Tokens {\cite{darcet2023vision}}. This recycled latent information restores spatial awareness without sacrificing efficiency.

\begin{table*}[t]
    \caption{Details of the seven real-world tasks and where we placed the cameras.}
    \label{table:realworld_tasks}
    \centering
    \renewcommand{\arraystretch}{1.15} 
    \begin{tabular*}{\textwidth}{@{\extracolsep{\fill}} l p{8.5cm} l}
        \toprule
        \textbf{Task Name} & \textbf{Task Instruction} & \textbf{Camera View} \\
        \midrule
        \textbf{Pick and Place} & \textit{Pick up the red cube and place it on the pallet} & Top, Side \\
        \textbf{Stack by Size} & \textit{Pick up the red cubes and stack them on the fixed blue platform in order from largest to smallest...} & Top, Side \\
        \textbf{Pull and Place (Jenga)} & \textit{Pull out the protruding blue Jenga block from the tower carefully, and place it on top of the stack} & Top, Wrist \\
        \textbf{Build Domino Line*} & \textit{Pick up the red, orange, yellow, and green blocks in that order, and place them upright in a straight line...} & Top, Wrist \\
        \textbf{Close Drawer} & \textit{Close the top drawer of the cabinet} & Top, Wrist \\
        \textbf{Move Bowl} & \textit{Pick up the silver bowl on the box and place it on the plate} & Top, Wrist \\
        \textbf{Clean Marker on Mirror**} & \textit{Pick up the eraser using mirror reflection, erase drawing from mirror} & Wrist \\
        \bottomrule
    \end{tabular*}
    \vspace{0.2em}
    \footnotesize
    \begin{minipage}{\textwidth}
    \vspace{0.2em}
        \textit{* We design a long-horizon manipulation task comprising an average of 900 frames per episode, which is 2–3 times longer than typical real-world tasks.}\\
        \textit{** To minimize the use of visual inputs, we restrict the agent to a single wrist-mounted camera and introduce a mirror as an auxiliary visual modality to compensate for the limited viewpoint.}
    \end{minipage}
\end{table*}

\subsection{Artifacts and Register Tokens in Vision Transformers}

Large ViTs~\cite{dosovitskiy2020image} use background image patches as scratchpads for global scene information, which supports learning but hurts local feature predictions~\cite{simeoni2021localizing, zheng2021rethinking, ranftl2021vision}. Learnable Register Tokens~\cite{darcet2023vision} address this issue by absorbing those artifacts. Although they are typically discarded after processing, we argue that they contain a useful spatial summary. RetoVLA uses these discarded tokens directly to guide robot motion.

\section{METHOD}

This section describes the RetoVLA architecture (Fig.~\ref{fig:architecture}). The core idea is to repurpose Register Tokens~\cite{darcet2023vision} as providers of spatial context. Instead of removing information, we recycle latent representations and route them directly into the Action Expert to provide geometric cues for motion planning.

\subsection{Architecture Overview and Information Flow}

RetoVLA retains the baseline structure but changes the internal data flow. Instead of sending only local patch features, we pass two streams. The Action Expert receives standard image patches together with Register Tokens \cite{darcet2023vision} that carry a global scene summary. A cross-attention layer combines these streams.

\subsection{Depth-Adaptive VLM Backbone}

To balance speed and capability, we use the first $N=L/2$ layers of the pre-trained VLM. SmolVLA~{\cite{shukor2025smolvla}} showed that this truncation accelerates inference while preserving semantic capability.

\subsection{Spatial Context Injection via Register Tokens}

We inject Register Tokens {\cite{darcet2023vision}} directly into the Action Expert in three steps.

\paragraph{Register Token Generation}
First, VLM image patch features ($\mathbf{P} \in \mathbb{R}^{B \times N \times D_{\text{vlm}}}$) enter a Spatial Context Aggregator, implemented as a standard multi-head attention block~\cite{vaswani2017attention}. Initial Register Tokens {\cite{darcet2023vision}} ($\mathbf{R}_{\text{init}} \in \mathbb{R}^{K \times D_{\text{vlm}}}$) act as the query, while image patches act as keys and values. This produces a global scene summary, $\mathbf{R}_{\text{scene}}$:
\begin{equation}
\mathbf{R}_{\text{scene}} = \text{Attention}(\mathbf{Q}=\mathbf{R}_{\text{init}}, \mathbf{K}=\mathbf{P}, \mathbf{V}=\mathbf{P})
\label{eq:aggregator}
\end{equation}
where $K$ is the number of Register Tokens.

\paragraph{Injection into the Action Expert}
We project $\mathbf{R}_{\text{scene}}$ to match the Action Expert and form key ($\mathbf{K}_{\text{reg}}$) and value ($\mathbf{V}_{\text{reg}}$) pairs. Concatenating these with the standard VLM pairs ($\mathbf{K}_{\text{vlm}}, \mathbf{V}_{\text{vlm}}$) lets the Action Expert access both local details and global context:
\begin{align}
\mathbf{K}_{\text{final}} &= \text{Concat}(\mathbf{K}_{\text{vlm}}, \sigma(g) \cdot \mathbf{K}_{\text{reg}}) \label{eq:key_injection} \\
\mathbf{V}_{\text{final}} &= \text{Concat}(\mathbf{V}_{\text{vlm}}, \sigma(g) \cdot \mathbf{V}_{\text{reg}}) \label{eq:value_injection}
\end{align}

\paragraph{Gating Mechanism}
Because global context can distract the policy during precision tasks, we introduce a learnable gate parameter $g$, passed through a sigmoid $\sigma$, to control the influence of the Register Tokens. This allows the model to adaptively balance local precision and global context.

\subsection{Training Objective: Conditional Flow Matching}

We train RetoVLA using conditional flow matching~\cite{lipman2022flow} to map pure noise to robot actions, conditioned on image and text inputs. Let $\mathbf{a}_0$ be the real robot action and $\mathbf{a}_1 \sim \mathcal{N}(0, \mathbf{I})$ be random noise. We define $\mathbf{a}_t = (1-t)\mathbf{a}_0 + t\mathbf{a}_1$ for $t \in [0, 1]$. The target vector is $\mathbf{u}_t = \mathbf{a}_1 - \mathbf{a}_0$. 

The model $\theta$ predicts this vector as $\mathbf{v}_\theta(\mathbf{a}_t, t, c)$, optimized via MSE:

\begin{figure}[t!]
    \centering
    \includegraphics[width=\columnwidth]{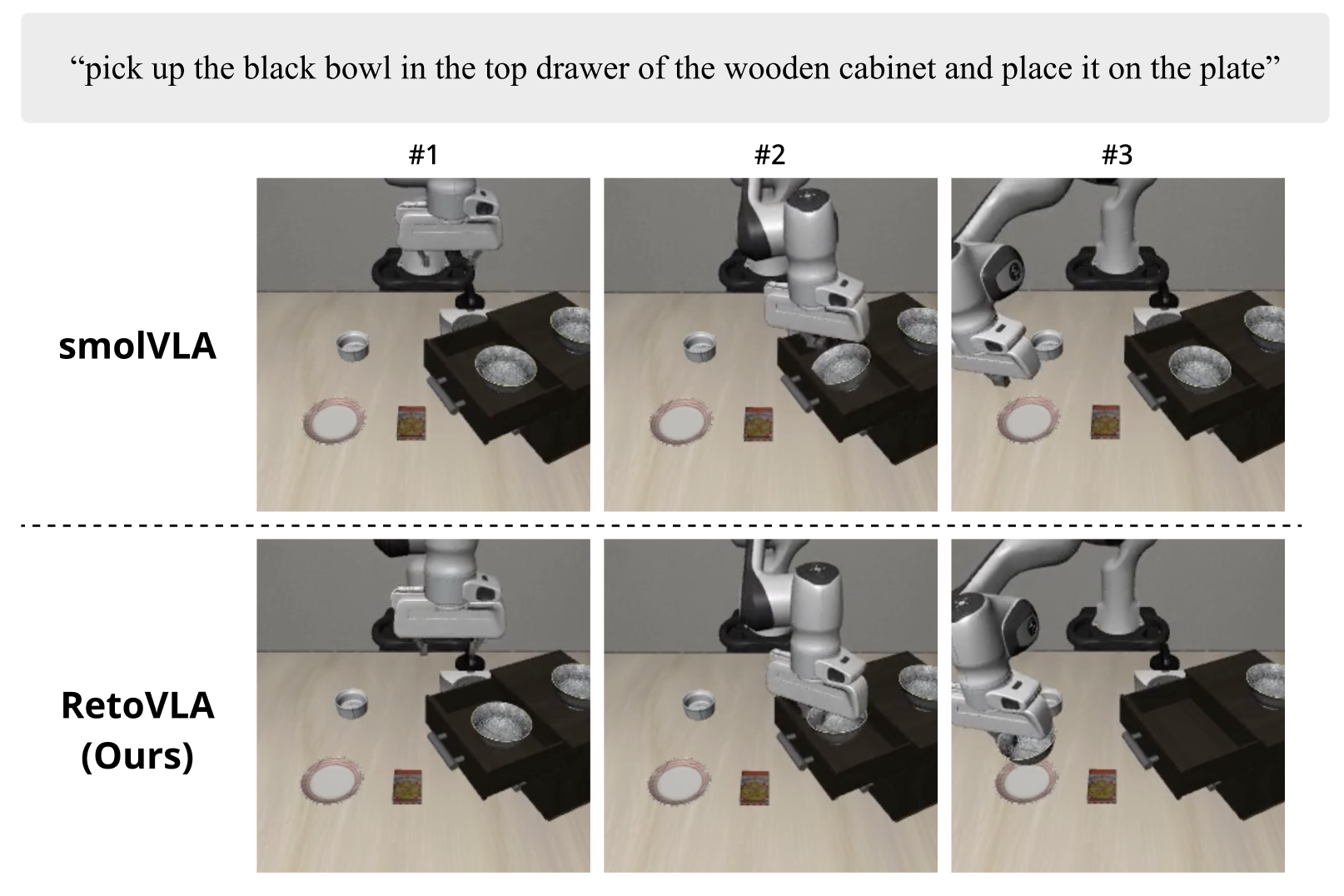}
    \caption{Reusing Register Tokens {\cite{darcet2023vision}} enables complex 3D spatial reasoning. The baseline SmolVLA {\cite{shukor2025smolvla}} grasps a visually similar but incorrect object, whereas RetoVLA correctly interprets the instruction ``in the top drawer'' by using the injected spatial context. This example highlights the model's ability to handle complex multi-step manipulation commands.}
    \label{fig:sim}
\end{figure}

\begin{equation}
\mathcal{L}_{\text{FM}} = \mathbb{E}_{t, \mathbf{a}_1, \mathbf{a}_0, c} \left[ \left\| \mathbf{v}_\theta(\mathbf{a}_t, t, c) - (\mathbf{a}_1 - \mathbf{a}_0) \right\|^2 \right]
\label{eq:flow_matching_loss}
\end{equation}

This objective trains the Action Expert to recover the correct action using spatial cues from the Register Tokens {\cite{darcet2023vision}}.

\begin{table}[b!]
\caption{Overall success rates on the four main categories of the LIBERO benchmark}
\label{tab:libero_summary}
\centering
\renewcommand{\arraystretch}{1.15} 
\renewcommand{\theadfont}{\bfseries}
\begin{tabular}{l c c c}
    \toprule
    \thead{Category} & \thead{Task \\ Characteristics} & \thead{SmolVLA {\cite{shukor2025smolvla}} \\ (SR)} & \thead{RetoVLA \\ (SR)} \\
    \midrule
    Spatial & Single spatial relations & 75.8\% & \textbf{76.2\%} \\
    \addlinespace 
    Object & \makecell[l]{Object-centric, \\ local manipulation} & 70.8\% & \textbf{71.8\%} \\
    \addlinespace 
    Goal & \makecell[l]{Goal-directed, \\ global placement} & 80.4\% & 80.4\% \\
    \addlinespace 
    10 (Long) & \makecell[l]{Long-horizon, \\ complex scenes} & 50.4\% & 50.4\% \\
    \bottomrule
\end{tabular}
\end{table}

\section{EXPERIMENTS}

We evaluated RetoVLA on the LIBERO benchmark, a real robot arm, and a custom simulation environment (Fig.~\ref{fig:tasks}).

\subsection{Experimental Setup}

\paragraph{Standardized Benchmark}
We used the `Spatial', `Object', `Goal', and `10 (Long)' task groups from the LIBERO benchmark~\cite{liu2023libero, libero-hf-dataset} to assess diverse skills.

\paragraph{Real-World Environment and Task Design}
We evaluated seven tasks on a custom 7-DOF robot arm (Table~\ref{table:realworld_tasks}), ranging from basic pick-and-place to long-horizon planning (`Build Domino Line') and 3D understanding (`Close Drawer'). We collected 1,804 human demonstrations to train both models on our custom hardware.

\begin{figure}[t!]
    \centering
    \includegraphics[width=\columnwidth]{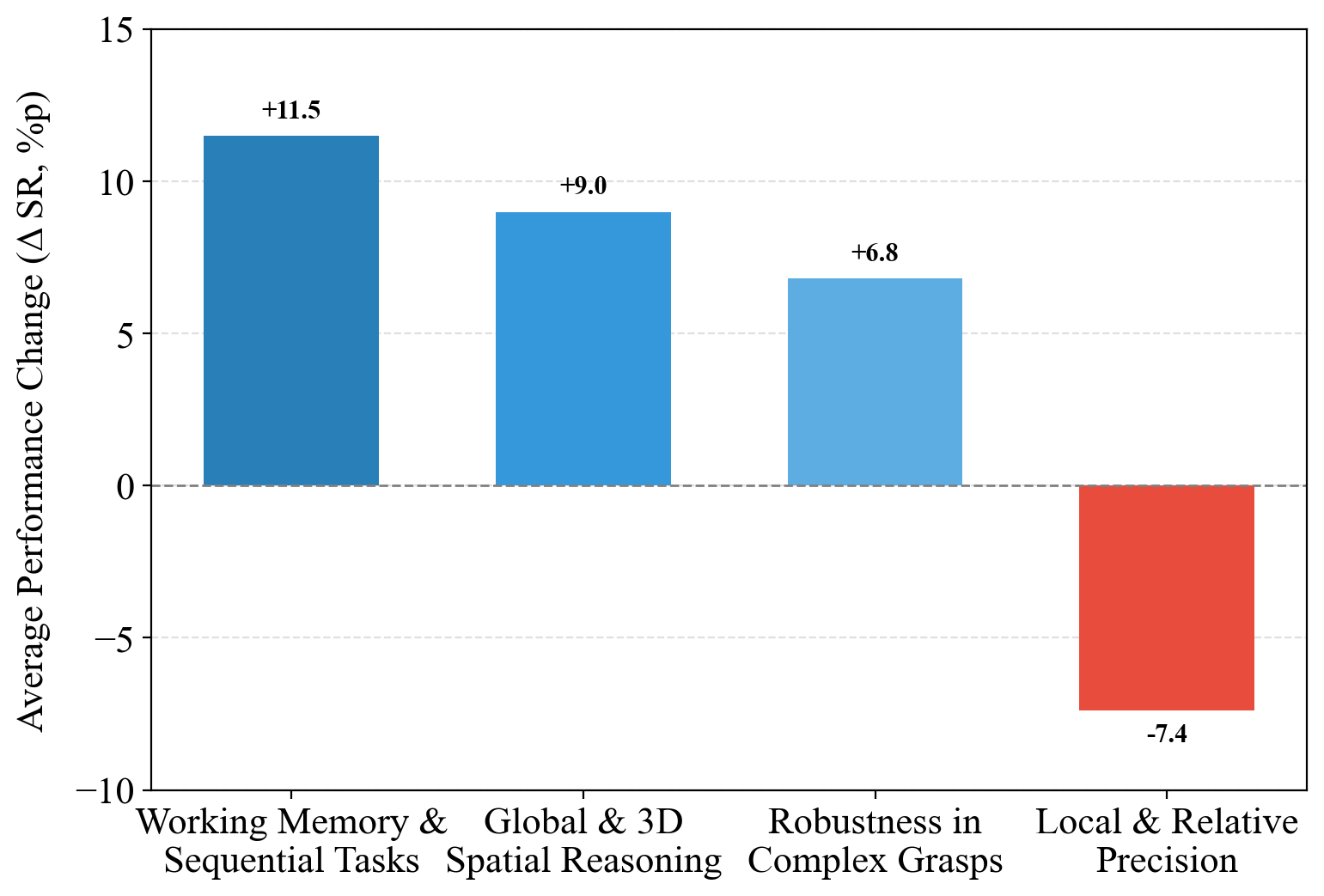}
    \caption{Breakdown of RetoVLA's performance by skill type. We plot the change in success rate relative to the baseline across all 40 LIBERO tasks. RetoVLA performs better on tasks that require memory and 3D understanding, but it loses some accuracy on tasks that demand extreme local precision.}
    \label{fig:core_capability}
\end{figure}

\paragraph{Custom Simulation Environment}
To evaluate the policy without physical noise or lighting shifts, we digitally replicated our real-world setup using Unity's MuJoCo engine.

\paragraph{Model and Training Setup}
For efficiency, we retained only the first 16 layers of SmolVLM2-500M~\cite{marafioti2025smolvlm}. Training lasted 100k steps with a batch size of 64. RetoVLA used two Register Tokens {\cite{darcet2023vision}} in its Action Expert; the baseline used none.

\begin{table}[t!]
    \caption{Success rates on Simulation Environment manipulation tasks}
    \label{tab:sim_success_rates}
    \centering
    \small  
    \renewcommand{\arraystretch}{1.15} 
    \renewcommand{\theadfont}{\bfseries} 
    \begin{tabular*}{\columnwidth}{@{\extracolsep{\fill}} l c c c} 
        \toprule
        \thead[l]{Task Name} & \thead{SmolVLA {\cite{shukor2025smolvla}} \\ (SR)} & \thead{RetoVLA \\ (SR)} & \thead{Performance \\ Change ($\Delta$)} \\
        \midrule
        Pick and Place & 88\%  & \textbf{96\%}  & \textbf{+6.0\%p} \\
        \addlinespace
        Stack by Size  & 86\%  & \textbf{88\%}  & \textbf{+2.0\%p} \\
        \addlinespace
        \makecell[l]{Pull and Place \\ (Jenga)} & 66\%  & \textbf{82\%}  & \textbf{+16.0\%p} \\
        \addlinespace
        \makecell[l]{Build Domino \\ Line}      & 28\%  & \textbf{52\%}  & \textbf{+24.0\%p} \\
        \addlinespace
        \makecell[l]{Clean Marker \\ on Mirror} & 46\%  & \textbf{56\%}  & \textbf{+10.0\%p} \\
        \addlinespace
        \midrule
        \textbf{MSR} & \makecell{62.8\% \\ $\pm$ 11.56\%} & \textbf{\makecell{74.8\% \\ $\pm$ 8.8\%}} & \textbf{+12.0\%p} \\
        \bottomrule
    \end{tabular*}
\end{table}

\begin{figure*}[t!]
    \centering
    \includegraphics[width=\textwidth]{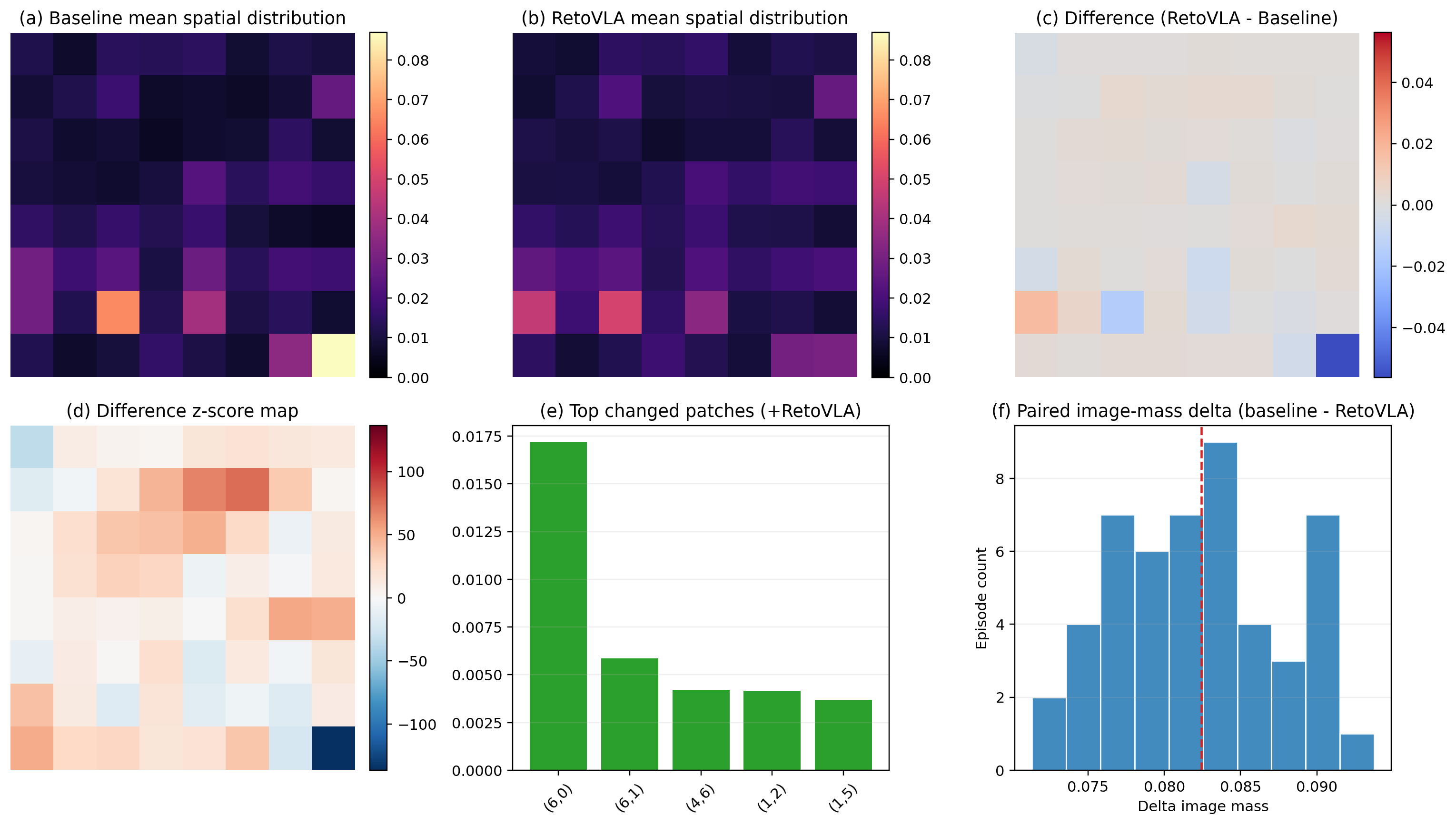}
    \caption{Visualization of efficient attention redistribution. (a, b) The mean spatial attention distributions show where the model is looking. (c, d) The difference maps (RetoVLA minus Baseline) reveal a fascinating pattern: RetoVLA actively ignores the broad, featureless background regions (shown in blue). It offloads this global context processing to the Register Tokens. (e) Instead, it sharpens its focus. The top increased patches (green bars) correspond precisely to the grippers and the target objects. (f) The consistent positive shift in image-mass delta confirms that RetoVLA relies less on raw image tokens overall, trusting the injected tokens to handle the ``big picture."}
    \label{fig:attention_map}
\end{figure*}

\subsection{LIBERO Benchmark Results}
While the overall scores in Table~\ref{tab:libero_summary} improve only slightly, Figure~\ref{fig:core_capability} reveals clear gains in Working Memory (+11.5\%p) and Global \& 3D Spatial Reasoning (+9.0\%p). These results suggest that token reuse improves 3D understanding. Figure~\ref{fig:sim} shows that RetoVLA opens the correct drawer by capturing room layout, unlike the baseline. A minor drop on tasks that require extreme local precision suggests that broad scene context can occasionally interfere with fine control (see Appendix and Fig.~\ref{fig:attention_map}).

\begin{table}[t!]
    \caption{Success rates on Real-World manipulation tasks}
    \label{tab:realworld_results}
    \centering
    \small  
    \renewcommand{\arraystretch}{1.15} 
    \renewcommand{\theadfont}{\bfseries} 
    \begin{tabular*}{\columnwidth}{@{\extracolsep{\fill}} l c c c} 
        \toprule
        \thead[l]{Task Name} & \thead{SmolVLA {\cite{shukor2025smolvla}} \\ (SR)} & \thead{RetoVLA \\ (SR)} & \thead{Performance \\ Change ($\Delta$)} \\
        \midrule
        Pick and Place & 86\%  & \textbf{92\%}  & \textbf{+6.0\%p} \\
        \addlinespace
        Stack by Size  & \textbf{80\%}  & 76\%  & -4.0\%p \\
        \addlinespace
        \makecell[l]{Pull and Place \\ (Jenga)} & 60\%  & \textbf{78\%}  & \textbf{+18.0\%p} \\
        \addlinespace
        \makecell[l]{Build Domino \\ Line}      & 12\%  & \textbf{40\%}  & \textbf{+28.0\%p} \\
        \addlinespace
        \makecell[l]{Clean Marker \\ on Mirror} & 38\%  & \textbf{52\%}  & \textbf{+14.0\%p} \\
        \addlinespace
        Close Drawer           & 60\%  & \textbf{96\%} & \textbf{+36.0\%p} \\
        \addlinespace
        Move Bowl             & 16\%  & \textbf{38\%}  & \textbf{+14.0\%p} \\
        \midrule
        \textbf{MSR} & \makecell{50.28\% \\ $\pm$ 11.06\%} & \textbf{\makecell{67.42\% \\ $\pm$ 9.07\%}} & \textbf{+17.14\%p} \\
        \bottomrule
    \end{tabular*}
\end{table}

\subsection{Real-World Experiments}
Real-world experiments (Table~\ref{tab:realworld_results}) show clear improvements. The mean success rate increases by 17.14\%p, from 50.28\% to 67.42\%. RetoVLA performs especially well on tasks that require deeper spatial understanding, such as `Build Domino Line' (+28\%p) and `Close Drawer' (+36\%p). The improvement on `Jenga' (+18\%p) further indicates that spatial context helps careful object interaction.

\subsection{Custom Simulation Experiments}
To isolate the impact of token reuse from physical noise, we conducted custom simulation experiments. Table~\ref{tab:sim_success_rates} shows an MSR gain of 12.0\%p, from 62.8\% to 74.8\%, with the largest improvements on `Build Domino Line' (+24.0\%p) and `Jenga' (+16.0\%p). The consistency across simulation, LIBERO, and real-world evaluation suggests that injecting Register Tokens {\cite{darcet2023vision}} improves spatial task performance.

\subsection{Analytical Studies: Understanding Attention and Causality}
Attention maps (Fig.~\ref{fig:register_analysis}) show that the model actively uses Register Tokens. They receive high attention weights across tasks (Fig.~\ref{fig:register_analysis}a, b). Sweeping the gate value ($g$) changes the predicted action (Fig.~\ref{fig:register_analysis}c), and replacing the tokens with noise reduces success rates (Fig.~\ref{fig:register_analysis}d). These results indicate that the tokens contain meaningful spatial information.

By relying on Register Tokens for global context, the model frees visual attention for task-relevant regions. Figure~\ref{fig:attention_map} shows a clear reduction in attention on flat background regions (blue areas in c, d). The saved attention shifts toward grippers and target objects (green bars in e), which helps explain the observed performance gains.

\section{CONCLUSIONS}

RetoVLA reuses Register Tokens {\cite{darcet2023vision}} to provide spatial context for robotic action generation, improving the performance of lightweight models on complex tasks.

During evaluation, RetoVLA remains relatively robust to moving shadows, likely because Register Tokens capture broad layout information and reduce sensitivity to lighting changes. However, highly reflective objects remain challenging, which indicates that complex texture perception is still difficult for small models.

Our method also has limitations. Performance drops slightly on tasks that require extreme local precision, which suggests the need for a more selective gating mechanism. In addition, we evaluate the approach only on a small model. Future work should test the same idea on larger backbones such as OpenVLA {\cite{kim2024openvla}} and on other robotic platforms, including mobile robots.

We will share our code, model weights, data, and hardware designs to support further research.

\section*{ACKNOWLEDGMENT}

This work was supported by the Future Challenge Defense Technology Research and Development Program through the Agency For Defense Development(ADD) grant funded by the Defense Acquisition Program Administration(DAPA) in 2025 (No.915134201) and by Gachon University research fund (GCU-202500670001).


\bibliographystyle{IEEEtran}
\bibliography{reference}

\section*{APPENDIX}

\subsection{Ablation Study}

\begin{table}[h!]
    \centering
    \caption{Ablation study on the number of Register Tokens {\cite{darcet2023vision}}.}
    \label{tab:ablation_registers}
    \renewcommand{\arraystretch}{1.15} 
    \begin{tabular*}{\columnwidth}{@{\extracolsep{\fill}} l c c}
        \toprule
        \textbf{Model Architecture} & \textbf{\# Register Tokens {\cite{darcet2023vision}}} & \textbf{Peak (SR)} \\
        \midrule
        SmolVLA {\cite{shukor2025smolvla}} (Baseline) & 0 & 75.8\% \\
        \midrule
        RetoVLA (Ours) & \textbf{2} & \textbf{76.2\%} \\
        RetoVLA (Ours) & 4 & 75.2\% \\
        RetoVLA (Ours) & 12 & 74.0\% \\
        RetoVLA (Ours) & 16 & 74.6\% \\
        \bottomrule
    \end{tabular*}
\end{table}

Table~\ref{tab:ablation_registers} shows that two Register Tokens are optimal on the LIBERO Spatial benchmark (76.2\% vs. 75.8\% for the baseline). Adding more tokens ($\ge 4$) degrades performance, likely because excessive global tokens act as noise and overwrite important local details. A small number of tokens provides the best balance between global context and local precision.

\subsection{Detailed LIBERO Benchmark Results}

Tables~\ref{tab:appendix_spatial}-\ref{tab:appendix_10} compare RetoVLA and the baseline at their best checkpoints (40k--50k for SmolVLA and 60k--70k for RetoVLA). SmolVLA often overfits after 50k steps, whereas RetoVLA remains stable and improves, indicating Register Tokens aid training stability. These results confirm RetoVLA excels in spatial memory tasks with minimal local precision trade-offs.

\subsection{Formal Definition of Analytical Metrics}

We define \textit{Attention Mass} ($M$) as the sum of cross-attention weights for a specific token group. Let $\alpha_{i,j}^{(l,h)}$ be the weight from Action Expert query $i$ to Vision Encoder key $j$ at layer $l$ and head $h$. Due to softmax, $\sum_{j} \alpha_{i,j}^{(l,h)} = 1$.

Register Mass ($M_{\text{reg}}$) and Image Mass ($M_{\text{img}}$) are the average attention across all queries ($N_q$), heads ($H$), and layers ($L$):
\begin{align}
M_{\text{reg}} &= \frac{1}{N_q H L} \sum_{l=1}^{L} \sum_{h=1}^{H} \sum_{i=1}^{N_q} \sum_{j \in \mathcal{K}_{\text{reg}}} \alpha_{i,j}^{(l,h)} \\
M_{\text{img}} &= \frac{1}{N_q H L} \sum_{l=1}^{L} \sum_{h=1}^{H} \sum_{i=1}^{N_q} \sum_{j \in \mathcal{K}_{\text{img}}} \alpha_{i,j}^{(l,h)}
\end{align}
where $\mathcal{K}_{\text{reg}}$ and $\mathcal{K}_{\text{img}}$ are the index sets for Register and Image Tokens.

To analyze visual focus shifts (Fig.~\ref{fig:attention_map}f), we calculate \textit{Delta Image Mass}:
\begin{equation}
\Delta M_{\text{img}} = M_{\text{img}}^{(\text{Baseline})} - M_{\text{img}}^{(\text{RetoVLA})}
\end{equation}
A positive $\Delta M_{\text{img}}$ means RetoVLA pays less attention to normal patches, successfully offloading broad background understanding to Register Tokens.

\textit{Action Delta} (Fig.~\ref{fig:register_analysis}c, d) is the $L_2$ distance between the normal prediction $\mathbf{v}_t$ and the changed prediction $\mathbf{\hat{v}}_t$ (e.g., when adding noise):
\begin{equation}
\Delta A = \|\mathbf{v}_t - \mathbf{\hat{v}}_t\|_2
\end{equation}

\begin{table}[h!]
    \centering
    \caption{Detailed comparison on the LIBERO Spatial benchmark.}
    \label{tab:appendix_spatial}
    \renewcommand{\arraystretch}{1.15}
    \renewcommand{\theadfont}{\bfseries}
    \begin{tabular*}{\columnwidth}{@{\extracolsep{\fill}} c c c c}
        \toprule
        \thead[l]{Index} & \thead{SmolVLA \\ (SR)} & \thead{RetoVLA \\ (SR)} & \thead{Change \\ ($\Delta$)} \\
        \midrule
        0 & \textbf{88.0\%} & 80.0\% & -8.0\%p \\
        1 & \textbf{88.0\%} & 80.0\% & -8.0\%p \\
        2 & 92.0\% & \textbf{98.0\%} & \textbf{+6.0\%p} \\  
        3 & 70.0\% & \textbf{86.0\%} & \textbf{+16.0\%p} \\ 
        4 & 50.0\% & \textbf{62.0\%} & \textbf{+12.0\%p} \\ 
        5 & \textbf{84.0\%} & 66.0\% & -18.0\%p \\
        6 & 92.0\% & \textbf{96.0\%} & \textbf{+4.0\%p} \\  
        7 & 62.0\% & \textbf{72.0\%} & \textbf{+10.0\%p} \\ 
        8 & \textbf{80.0\%} & 68.0\% & -12.0\%p \\
        9 & 52.0\% & \textbf{54.0\%} & \textbf{+2.0\%p} \\  
        \midrule
        \textbf{MSR} & 75.8\% & \textbf{76.2\%} & \textbf{+0.4\%p} \\ 
        \bottomrule
    \end{tabular*}
\end{table}

\begin{table}[h!]
    \centering
    \caption{Detailed comparison on the LIBERO Object benchmark.}
    \label{tab:appendix_object}
    \renewcommand{\arraystretch}{1.15}
    \renewcommand{\theadfont}{\bfseries}
    \begin{tabular*}{\columnwidth}{@{\extracolsep{\fill}} c c c c}
        \toprule
        \thead[l]{Index} & \thead{SmolVLA \\ (SR)} & \thead{RetoVLA \\ (SR)} & \thead{Change \\ ($\Delta$)} \\
        \midrule
        0 & 56.0\% & \textbf{58.0\%} & \textbf{+2.0\%p} \\  
        1 & \textbf{82.0\%} & 74.0\% & -8.0\%p \\
        2 & 70.0\% & \textbf{76.0\%} & \textbf{+6.0\%p} \\  
        3 & 50.0\% & 50.0\% & 0.0\%p \\
        4 & \textbf{94.0\%} & 82.0\% & -12.0\%p \\
        5 & 54.0\% & 54.0\% & 0.0\%p \\
        6 & 82.0\% & \textbf{84.0\%} & \textbf{+2.0\%p} \\  
        7 & 54.0\% & \textbf{64.0\%} & \textbf{+10.0\%p} \\ 
        8 & 78.0\% & \textbf{92.0\%} & \textbf{+14.0\%p} \\ 
        9 & \textbf{88.0\%} & 84.0\% & -4.0\%p \\
        \midrule
        \textbf{MSR} & 70.8\% & \textbf{71.8\%} & \textbf{+1.0\%p} \\ 
        \bottomrule
    \end{tabular*}
\end{table}

\begin{table}[h!]
    \centering
    \caption{Detailed comparison on the LIBERO Goal benchmark.}
    \label{tab:appendix_goal}
    \renewcommand{\arraystretch}{1.15}
    \renewcommand{\theadfont}{\bfseries}
    \begin{tabular*}{\columnwidth}{@{\extracolsep{\fill}} c c c c}
        \toprule
        \thead[l]{Index} & \thead{SmolVLA \\ (SR)} & \thead{RetoVLA \\ (SR)} & \thead{Change \\ ($\Delta$)} \\
        \midrule
        0 & 60.0\% & \textbf{74.0\%} & \textbf{+14.0\%p} \\ 
        1 & 94.0\% & \textbf{100.0\%} & \textbf{+6.0\%p} \\  
        2 & \textbf{86.0\%} & 84.0\% & -2.0\%p \\
        3 & \textbf{70.0\%} & 68.0\% & -2.0\%p \\
        4 & \textbf{90.0\%} & 86.0\% & -4.0\%p \\
        5 & \textbf{90.0\%} & 82.0\% & -8.0\%p \\
        6 & 54.0\% & \textbf{62.0\%} & \textbf{+8.0\%p} \\   
        7 & 92.0\% & 92.0\% & 0.0\%p \\
        8 & \textbf{94.0\%} & 84.0\% & -10.0\%p \\
        9 & \textbf{74.0\%} & 72.0\% & -2.0\%p \\
        \midrule
        \textbf{MSR} & 80.4\% & 80.4\% & 0.0\%p \\
        \bottomrule
    \end{tabular*}
\end{table}

\begin{table}[h!]
    \centering
    \caption{Detailed comparison on the LIBERO 10 benchmark.}
    \label{tab:appendix_10}
    \renewcommand{\arraystretch}{1.15}
    \renewcommand{\theadfont}{\bfseries}
    \begin{tabular*}{\columnwidth}{@{\extracolsep{\fill}} c c c c}
        \toprule
        \thead[l]{Index} & \thead{SmolVLA \\ (SR)} & \thead{RetoVLA \\ (SR)} & \thead{Change \\ ($\Delta$)} \\
        \midrule
        0 & \textbf{28.0\%} & 24.0\% & -4.0\%p \\
        1 & 38.0\% & \textbf{56.0\%} & \textbf{+18.0\%p} \\ 
        2 & 64.0\% & \textbf{74.0\%} & \textbf{+10.0\%p} \\ 
        3 & 76.0\% & \textbf{82.0\%} & \textbf{+6.0\%p} \\  
        4 & \textbf{32.0\%} & 24.0\% & -8.0\%p \\
        5 & 82.0\% & \textbf{94.0\%} & \textbf{+12.0\%p} \\ 
        6 & \textbf{36.0\%} & 26.0\% & -10.0\%p \\
        7 & 10.0\% & \textbf{22.0\%} & \textbf{+12.0\%p} \\ 
        8 & \textbf{62.0\%} & 52.0\% & -10.0\%p \\
        9 & \textbf{76.0\%} & 50.0\% & -26.0\%p \\
        \midrule
        \textbf{MSR} & 50.4\% & 50.4\% & 0.0\%p \\
        \bottomrule
    \end{tabular*}
\end{table}

\subsection{Detailed Causal Analysis of Register Tokens}

As outlined in the main text, we conducted a causal analysis to determine whether the tokens directly influence the robot's decisions (Fig.~\ref{fig:register_analysis}). Empirical results confirm the active utilization of these tokens, which consistently exhibit high attention weights across both successful and failed executions (a). By relying on these tokens for global context, the model correspondingly allocates less attention mass to standard image patches (b).

Furthermore, we investigated whether perturbing the tokens alters the generated actions. Sweeping the gate parameter ($g$) yields a direct shift in the predicted action (c), and substituting the tokens with random noise significantly degrades the success rate (d). These findings demonstrate that Register Tokens encode meaningful spatial representations rather than acting merely as empty noise absorbers.

\subsection{Detailed Visualization of Spatial Attention Shift}

Figure~\ref{fig:attention_map} illustrates the redistribution of visual attention within RetoVLA. The difference maps (c, d) reveal a distinct reduction in attention mass over featureless background regions, such as tables or walls (indicated in blue). This shift suggests that the model effectively offloads global context processing to the Register Tokens.

By conserving attention capacity in these background regions, the model can focus more precisely on task-relevant features. Panel (e) demonstrates that this reclaimed attention is redirected specifically toward the robot grippers and target objects. Finally, panel (f) confirms this trend across numerous episodes: RetoVLA exhibits a consistent decrease in its reliance on standard image patches for global scene comprehension, thereby enabling the Action Expert to concentrate on fine-grained manipulation.


\end{document}